\documentclass[letterpaper, 10 pt, conference]{IEEEconf}
\IEEEoverridecommandlockouts
\usepackage{cite}
\usepackage{amsmath,amssymb,amsfonts}
\usepackage{algorithmic}
\usepackage{graphicx}
\usepackage{textcomp}
\usepackage{xcolor}
\usepackage{verbatim}
\def\BibTeX{{\rm B\kern-.05em{\sc i\kern-.025em b}\kern-.08em
    T\kern-.1667em\lower.7ex\hbox{E}\kern-.125emX}}
\begin{document}

\title{Multi-Object Grasping -- Estimating the Number of Objects in a Robotic Grasp
}

\author{Tianze Chen, Adheesh Shenoy, Anzhelika Kolinko, Syed Shah and Yu Sun
\thanks{The authors are from the Robot Perception and Action Lab (RPAL) of Computer Science and Engineering Department, University of South Florida, Tampa, FL 33620, USA. Email: \texttt{\{tianzechen, shenoy, akolinko, mutahar,yusun\}@usf.edu}. Adheesh Shenoy and Syed Shah are undergraduate students.}}

\maketitle

\begin{abstract}
A human hand can grasp a desired number of objects at once from a pile based solely on tactile sensing. To do so, a robot needs to grasp within a pile, sense the number of objects in the grasp before lifting, and predict the number of objects that will remain in the grasp after lifting. It is a challenging problem because when making the prediction, the robotic hand is still in the pile and the objects in the grasp are not observable to vision systems. Moreover, some objects that are grasped by the hand before lifting from the pile may fall out of the grasp when the hand is lifted. This occurs because they were supported by other objects in the pile instead of the fingers of the hand. Therefore, a robotic hand should sense the number of objects in a grasp using its tactile sensors before lifting. This paper presents novel multi-object grasping analyzing methods for solving this problem. They include a grasp volume calculation, tactile force analysis, and a data-driven deep learning approach. The methods have been implemented on a Barrett hand and then evaluated in simulations and a real setup with a robotic system.  The evaluation results conclude that once the Barrett hand grasps multiple objects in the pile, the data-driven model can predict, before lifting, the number of objects that will remain in the hand after lifting. The root-mean-square errors for our approach are 0.74 for balls and 0.58 for cubes in simulations, and 1.06 for balls, and 1.45 for cubes in the real system.
\end{abstract}

\section{Introduction} 
Grasping multiple objects at once from a pile is a common task for us.  It is so common that we have the word ``handful'' to describe a quantity that fills the hand. As children, we grasped a handful of candies from a bowl.  When cooking, we pick a handful of Brussels sprouts from a bag. When we make a drink, we pick up two or three ice cubes at once from an ice bucket.  If we want multiple objects at once, we rarely pick them up one by one. Instead, we do so by picking all the objects at once because it is more efficient. 

A robot should gain the skill of grasping the desired number of objects from a pile.  Strategies for stable grasping for multiple objects are discussed in \cite{harada1998enveloping} and \cite{harada2002active}, however, the target objects are outside the bin and traditional grasp quality measures were used to analyze the grasps \cite{lin2015grasp,LinISRR2013,lin2015task,Sun2016}. Unfortunately, we haven't seen any reports of grasping a desired number of objects at once from a pile.
Grasping multiple objects from a pile is a difficult problem since estimating the poses of multiple objects in a pile is very challenging and prone to error \cite{agrawal2010vision}. Even advanced vision algorithms cannot reliably estimate their poses when they are occluded by other objects with the same color and texture. In certain situations, the object may be translucent and hard to detect by a vision system. For example, in the 2019 and 2020 IROS Robotic Grasping and Manipulation Competition (RGMC) \cite{sun2021research}, all teams failed at picking ice cubes from an ice bucket. Moreover, the contact between objects and the fingers and palm is complex and not fully observable. In addition to visual occlusion, we may also have tactile occlusion.  Some grasped objects may not have direct contact with the hand.  

When we grasp the desired number of objects from a pile, we not only sense the number of objects in the grasp but also need to predict the quantity that will remain in the grasp after we lift the hand.  If it is lower than the desired number, we open our fingers to get more before lifting the hand. 
Similar to a human hand, a robotic hand has to use tactile sensors, force sensors, torque sensors, or strain gauge sensors to predict the quantity grasped.  The tactile sensor's resolution is usually sparse, and the coverage is limited.  Therefore, it is very difficult to get a whole picture of how many objects are in hand.  Even more difficult is to predict how many will remain in the hand after lifting because some of them are only partially restrained by the fingers and partially supported by other objects in the pile. When the hand is lifting, objects may not have sufficient support and may fall out of the grasp.

Tactile sensing has been recognized as a critical perception component in object grasping and manipulation, mainly to mitigate the challenges caused by visual occlusions. 
It has been used with vision sensors to estimate the location of an object relative in a world coordinate system using a tactile sensor array \cite{chhatpar2005particle, chebotar2014learning}, embedded force sensors on a robotic hand \cite{corcoran2010measurement}, a 6-axis force/torques sensor on a robot's wrist \cite{petrovskaya2006bayesian, javdani2013efficient,bimbo2015global,petrovskaya2011global,saund2017touch}. In those works, tactile/force sensors are used to reduce the uncertainty in the perception through a sequence of touch actions.  

Tactile sensors have also been used with vision sensors to track an object's location in-hand for in-hand manipulation \cite{liang2020hand, pfanne2018fusing, alvarez2017tactile}.  Many works use contact locations to estimate the object's pose with Bayesian or particle filtering \cite{corcoran2010measurement, platt2011using,chalon2013online,zhang2013dynamic,vezzani2017memory, ding2018hand}. Several works fuse tactile features with hand poses, force/torque sensor reading, and visual features to improve the estimation accuracy \cite{hebert2011fusion,pfanne2018fusing}. 
Tactile sensing can also be used to estimate the object's geometry \cite{pezzementi2011object, luo2015novel}, material, surface texture, 
stiffness \cite{silvera2012interpretation, tawil2011touch,chitta2011tactile,russell2000object}, and help in learning manipulation skills \cite{lin2012learning,huang2019dataset, lin2015robot,lin2014grasp}.  A detailed review on technologies of perceiving object properties through robotic tactile sensing can be found at \cite{luo2017robotic}. 

However, most of the current works focus on estimating the location or property of a single object either before grasping or after grasping.  We have not found any work for estimating the number of objects in a grasp.  Therefore, this paper focuses on exploring and presenting the difficulties and opportunities in perceiving the number of objects in a grasp.
Due to the difficulty of obtaining the ground-truth poses of multiple objects in hand, this paper only focuses on estimating the number of objects, not the objects' poses.

We first present a simple analytical approach that uses tactile readings to estimate the number of objects in a grasp using gravity.  However, since this approach does not perform well before the hand is lifted from the pile, its application is limited.  We then present another approach that estimates the number of objects in a grasp based on the grasp volume.  We calculate the grasp volume based on the hand's posture.  The volume can estimate the maximum number of objects in a grasp reasonably accurately, but not the exact number. Finally, we present a data-driven deep learning approach that can produce a good estimation of the final number of objects that remain in the grasp after the hand is lifted from the pile. The estimation result will be used to guide the robotic hand to grasp the desired number of objects in a pile. 

\section{Methodology}
\subsection{Problem Statement} 
If a robotic hand can grasp a desired number of objects at once from a pile, the robot should estimate the number of objects in the grasp before the hand is lifted. However, when the hand is still in a pile, a vision system can not see how many objects are in the grasp because of occlusion.  The robot has to rely on the hand pose, tactile sensors on the fingers and palm, and the joints' torque sensors to figure out how many objects are in a solid grasp and will remain in the grasp after the hand is lifted.  

When the hand is still in a pile, several objects will be in contact with the fingers. Some of them are firmly grasped by the fingers and palm, while others may loosely lie on a finger and fall once the hand is lifted.  Therefore, it is challenging to estimate the number of objects that will remain grasped after the hand is lifted.   

This paper uses a Barrett hand in the analysis and evaluation.  The presented approaches can be generalized to other robotic hands with tactile sensors. 

\subsection{Sensing with Gravity Force}
\subsubsection{After Lifting}
When objects are held by the robotic hand outside the bin, the wrist force sensor can sense the gravity of the objects and use it to estimate the number of objects grasped given the weight of a single object. If a robotic system doesn't have a precise wrist sensor, the contact and force from the tactile sensors can be used to estimate the gravity forces on the objects in hand and derive the number of the objects. For example, our evaluation system has a UR5e robotic arm with a wrist force sensor with a resolution of $2.5N$ and an accuracy variance of $4N$.  The accuracy variance is about $50-200$ times of our targeted objects: the ping-pong balls, foam cubes, and fortune cookies. Therefore, we cannot depend on the readings from this sensor and need to use the tactile sensors on the hand.

\subsubsection{Before lifting}
Even with an accurate force sensor, when the hand is in the pile, the wrist force sensor cannot isolate the object's gravity force from the hand's contact force with the remaining pile. Besides, before lifting, some objects could be supported by both a finger and another object.  Some of them may fall out of the hand when it is lifted.  Instead of estimating the number of objects directly based on the weight, we train a linear regression model that relates the number of objects in grasp with the contact force reading's vertical components.

\subsection{Grasp Volume Model}

We can roughly estimate the number of objects that could be grasped based on the volume that the grasp forms and the volume of the object. A grasp virtually separates the space into in-grasp space, out-grasp space, and boundary. The objects in the in-grasp space could be grasped. The objects on the boundary, that is, the space between the fingers, could also be grasped but have a higher chance of falling out. However, the objects in the out-grasp space cannot be grasped. Therefore, computing the volume of the in-grasp space and the boundary area could provide an estimation of the number of objects can be potentially grasped successfully.

\begin{figure}[htbp]

\centerline{

\includegraphics[height=4cm]{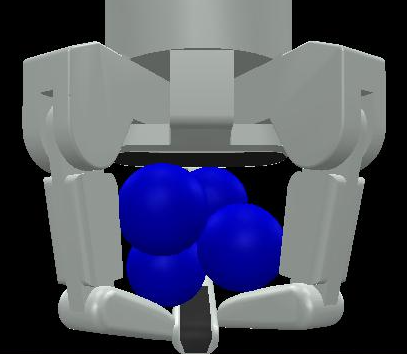}

\includegraphics[height=4cm]{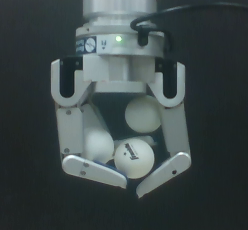}}

\caption{A Barrett Hand holds multiple spheres in both a simulation and a real environment. }

\label{fig-bh}

\end{figure}

\begin{figure}[htbp]
\begin{center}
\includegraphics[width=0.45\linewidth]{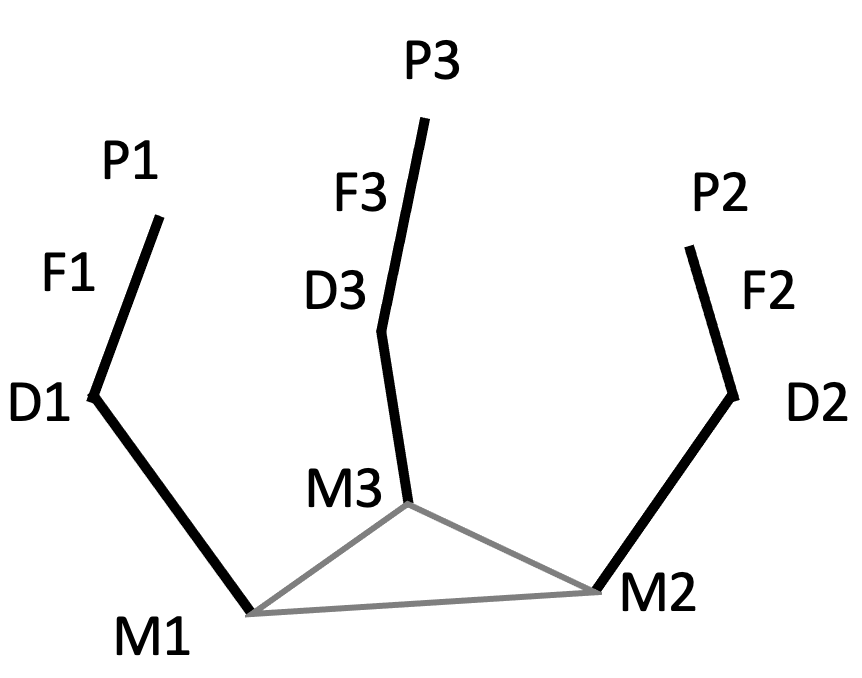}
\includegraphics[width=0.5\linewidth]{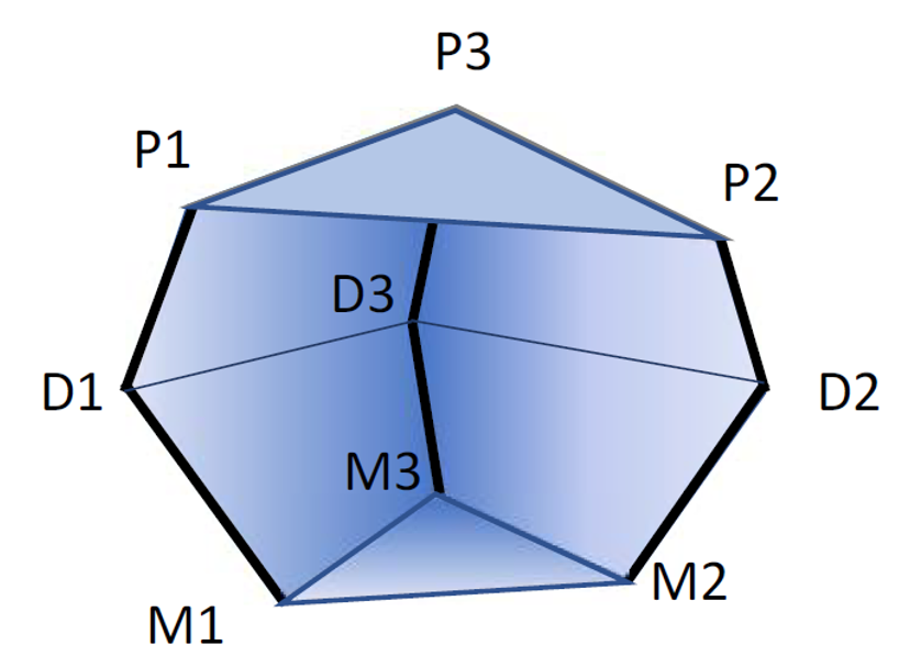}\\
(A)~~~~~~~~~~~~~~~~~~~~~~~~~~~~~(B)
\end{center}
\vspace{-4mm}
\caption{(A) Barrett Hand.
(B) The grasp space can be formed using a convex hull.}

\label{fig-bhand-model}

\end{figure}

We use a Barrett Hand (as shown in Figure \ref{fig-bh}) as an example to present our approach in calculating the volume of a grasp. As illustrated in Figure \ref{fig-bhand-model}, a space of a grasp formed by a Barrett Hand can be modeled with a polyhedron. M1 to M3 are the Metacarpophalangeal Joints, D1 to D3 are Distal interphalangeal joints, and P1 to P3 are the fingertip points. These coordinates can be computed using Barrett Hand's forward kinematic model. Naturally, treating each point on the hand as a coordinate, a convex hull similar to Figure \ref{fig-bhand-model} can be created to enclose the space within the boundary of the points. Using this convex hull, the volume of the in-grasp space can be generated for a given grasp. 

The number of objects that can fit into that volume depends on both the grasp volume and the object's geometry. Estimating how many objects can fit in an irregular shape formed by a hand is a well-known but extremely difficult packing problem. To compute a rough estimate, we calculate the object's maximum packing density when they are tightly packed in space and use it to calculate the potential number of objects by multiplying the grasp volume with the object's maximum packing density and dividing the result with the object's volume. Although the result is an overestimate, it provides a useful upper-bound number.

\subsection{Data-driven model}
\begin{figure}[]
    \centering
    \includegraphics[width=\linewidth]{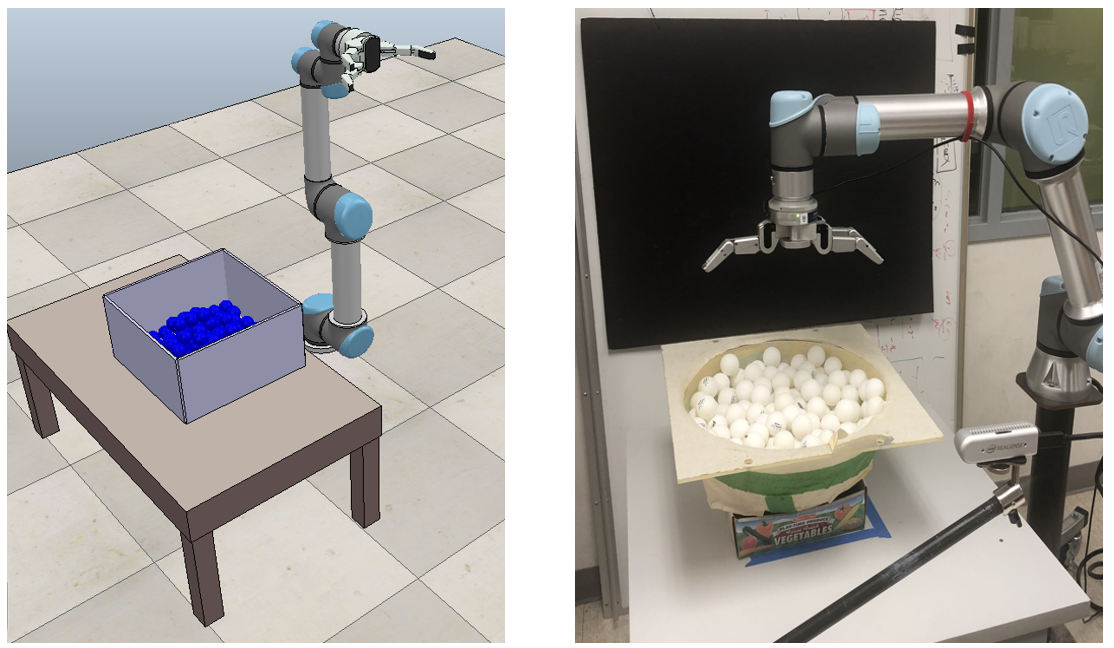}
    \caption{Setup of the system (left) in simulation and real world (right). The Barrett hand is attached to UR5e robot arm as end-effector. The objects to be grasped is in a box on the table in front of the robot.}
    \label{fig-system_setup}
\end{figure}

\subsubsection{Data collection}
Collecting grasping data from a real system is expensive since the chance of achieving a successful grasp (one or more objects in the grasp after the hand is lifted) is only around $50\%$. To make the data collection more efficient, we first collect data in simulation software and use the poses that are more likely to achieve a successful grasp to collect data in the real system.

\begin{figure}[htbp]
    \centering
    \includegraphics[width=0.8\linewidth]{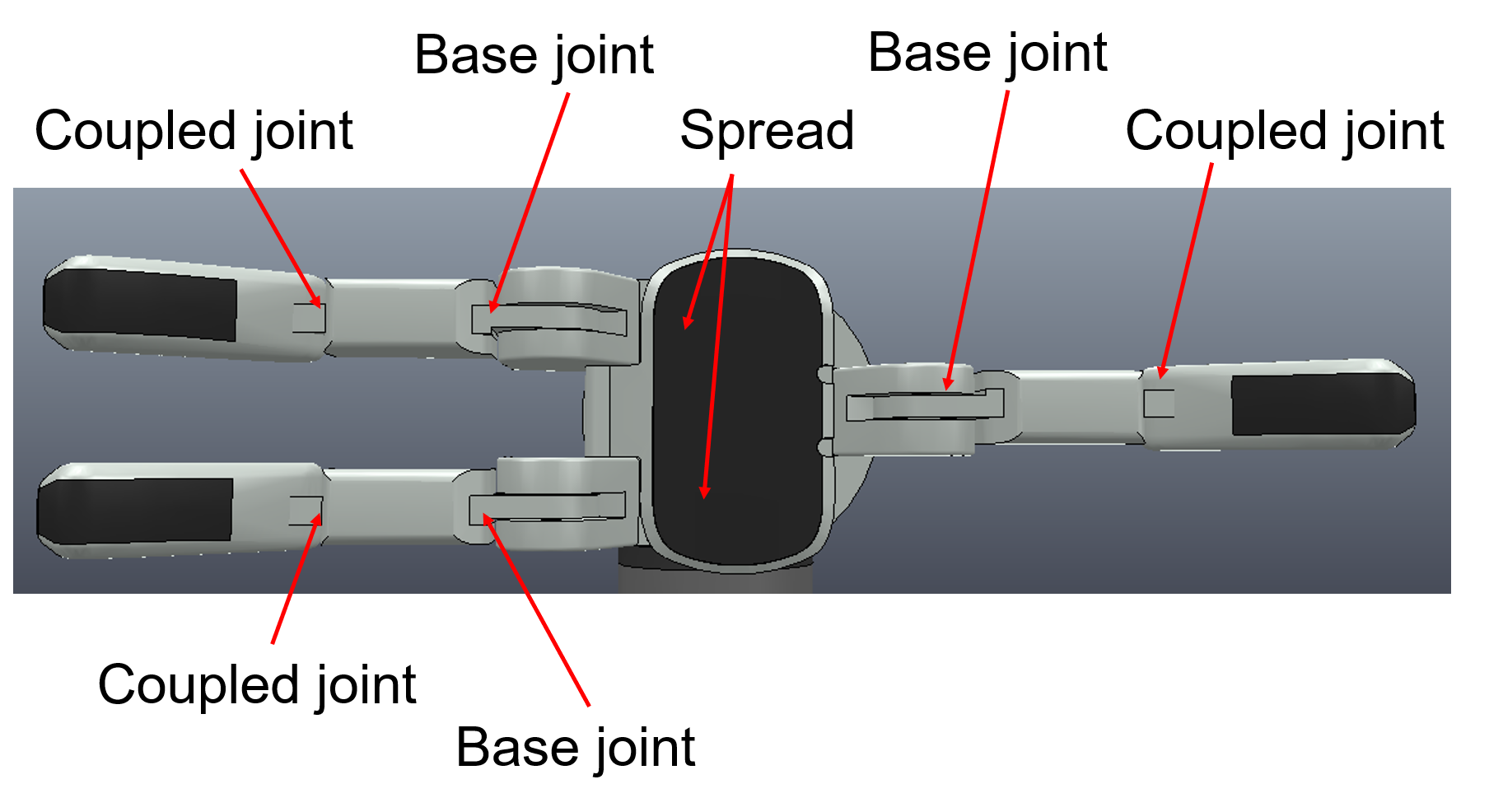}
    \vspace{-3mm}
    \caption{Joints on Barrett hand }
    \label{fig-Barrett_hand_joints}
\end{figure}

\paragraph{Real system setup}
The real system is composed of a UR5e robot arm, a Barrett hand, and a bowl of objects. The Barrett hand  has seven joints, as shown in Figure \ref{fig-Barrett_hand_joints}, and each hand pose contains the readings from all the joints. The palm and fingers each have $24$ tactile sensors.
Each finger also has a strain gauge sensor measuring the coupled joint torque.  We monitor the strain gauge readings during the data collection to avoid applying an extremely large force to the finger joints.

\begin{figure}[htbp]
    \centering
    \includegraphics[width=0.7\linewidth]{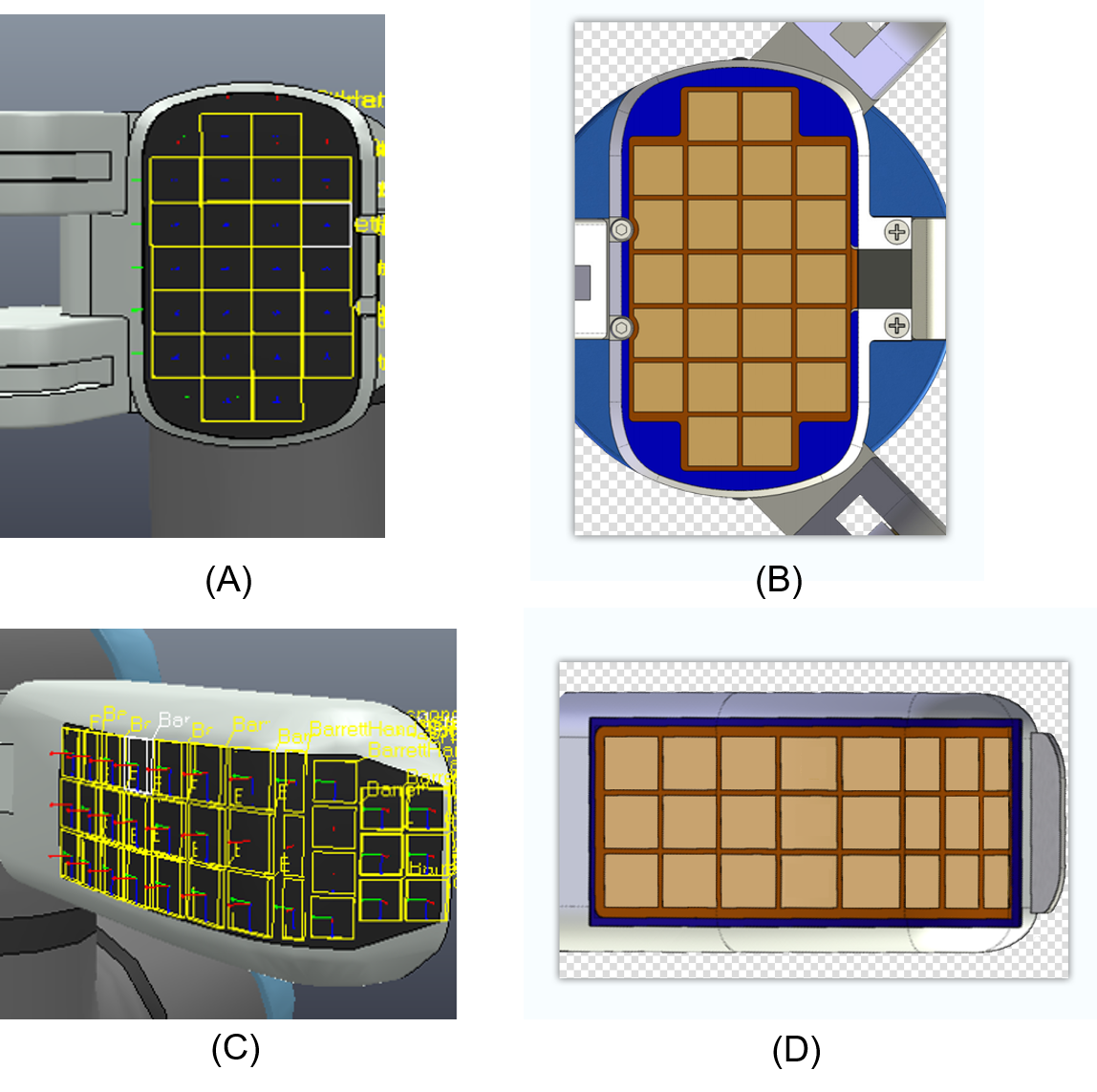}
    \vspace{-5mm}
    \caption{(A) and (C) shows the tactile sensors we manually attach in simulation on palm and fingertips respectively. (B) and (D) shows the tactile sensors distribution on the palm and fingertips respectively of Barrett hand in real system and they are provided by the manufacture of Barrett hand.}
    \label{fig-Tactile_sensors_comparison}
\end{figure}

\paragraph{Simulation setup}
We use the $Coppelia$ simulator to simulate the real system. We use an existing UR5 robotic arm.  We utilize the existing Barrett Hand model and embed tactile sensors based on the real Barrett hand. We attach $24$ tactile sensors onto the palm and $34$ tactile sensors onto each fingertip. A comparison of how the tactile sensors are distributed in simulation and the real system can be found in Figure \ref{fig-Tactile_sensors_comparison}. In order to have the same number of tactile readings in the similar location as the real Barrett hand, we will add the readings of the tactile sensor's neighbors and use the average value to represent the tactile sensor reading. 

\begin{figure}[htbp]
    \centering
    \includegraphics[width=0.5\textwidth]{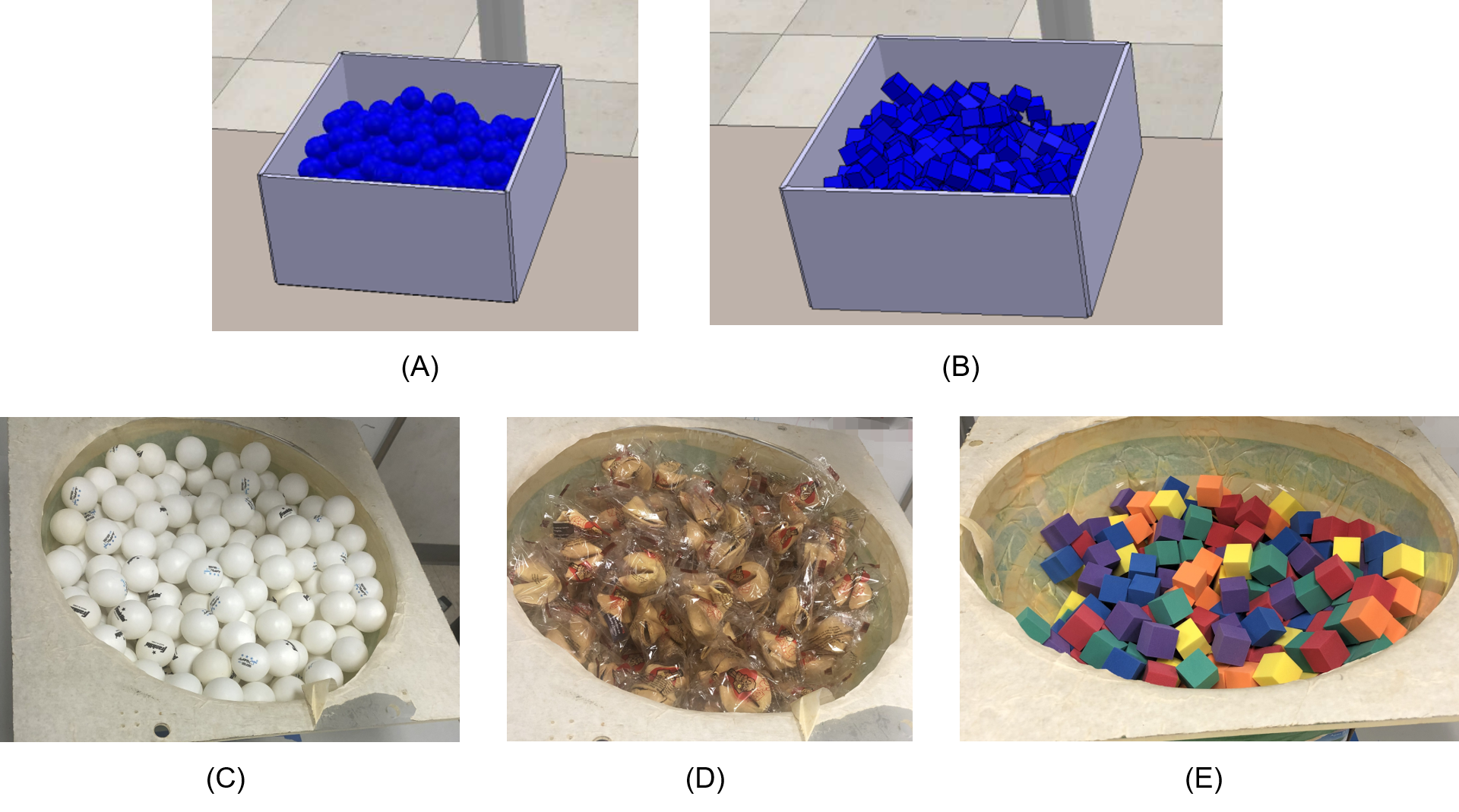}
    \caption{Objects we use for data collection: (A) and (B) are ping-pong ball and cube we use in simulation. (C) ping-pong ball, (D) fortune cookies and (E) foam cubes are the objects we use for data collection in real system.}
    \label{fig-object_picture}
\end{figure}

\paragraph{Objects}
The objects we use in the real system and the simulation are shown in Figure \ref{fig-object_picture}.

\paragraph{Data collection strategy}
We perform data collection on all potential pre-grasps in simulation. We choose the spread angle with the sampling step size of $20^{\circ}$ from range $0^{\circ}$ to $360^{\circ}$ and each finger angle with the sampling step size of $6^{\circ}$ from range $30^{\circ}$ to $90^{\circ}$. 

We remove the symmetric poses such that, in the end; we obtain a pre-grasp set with $9000$ elements. We repeat the grasping with each pre-grasps $10$ times in simulation on ping-pong balls and $6$ times in simulation on cubes to obtain the hand poses that have a relatively high grasp success rate. 
Since most of the grasps end with zero or one object, we choose $115$ pre-grasps from the $9000$ pre-grasp set for ping-pong balls and $130$ out of the $9000$ pre-grasp set for cubes because of their high success rate. We use the ping-pong ball's pose set to collect data in the real system by collecting $10$ trials for each ping-pong ball and $1$ trial for each on fortune cookies. We use the cube's pose set to collect data in the real system with foam cubes $1$ time.

\paragraph{Data formatting}
We record the number of objects grasped at the end of each trial. We also record the hand pose, the tactile sensor readings, and the strain gauge (coupled joints torque) readings when the hand is about to lift. Therefore, the size of each data point is $107$. 

\subsubsection{Deep learning models}
The sensing with force and the grasp volume calculation methods only consider one aspect that influences the grasping result. However, it is hard to develop a solution analytically, considering all the aspects that affect the final grasping result. 
To consider all modalities, we design a deep learning classifier to estimate the number of objects in a robotic grasp. 
The whole model can be represented as 
\begin{equation} \label{classifier}
n = f(\mathbf{h}, \mathbf{t}, \mathbf{s}),
\end{equation}
where $\mathbf{h}$ is a vector representing the Barrett hand pose, $\mathbf{t}$ is the vector representing the tactile reading array from all the tactile sensors, and $\mathbf{s}$ is the vector containing the three strain gauge readings of the coupled joints. The output $n$ is the prediction of the number of objects being grasped in the Barrett hand when the grasping trial has finished.

\begin{figure*}[htbp]
    \centering
    \includegraphics[width=0.95\textwidth]{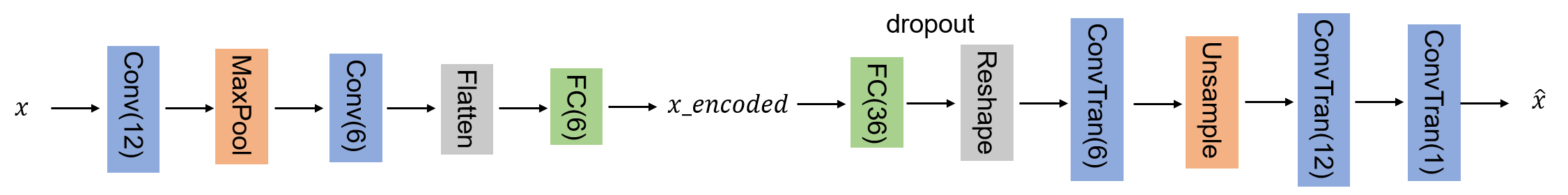}
    \vspace{-4mm}
    \caption{$x$ is the input tactile reading array with the size of $24$ on one of the four regions in Barrett hand, $x\_encoded$ is the encoded result of input tactile reading with size $6$ and $\hat{x}$ is the result from decoding with the same size as $x$. Conv(12) and Conv(6) refer to convolution layers with $12$ and $6$ filters. ConvTran(6), ConvTran(12) and ConvTran(1) refers to de-convolution layers with $6$, $12$ and $1$ filters. All the convolution and de-convolution layers have the same kernel size of $3$ by $3$ with a stride of $1$, and padding enabled. FC(6) and FC(36) refer to fully connected layers with $6$ and $36$ units. Max-pooling and Unsample layers have the same kernel size of $2$ by $2$. The dropout rate is 0.5.}
    \label{fig-auto-encoder}
\end{figure*}

\paragraph{Auto-encoder}
We have also implemented an auto-encoder to reduce the dimension of the tactile sensor input. The structure of the auto-encoder is shown in Figure \ref{fig-auto-encoder}. We train three auto-encoders for the tactile sensor readings from the palm, fixed finger, and two moving fingers, respectively. The output of the encoder has a dimension of $6$. We concatenate the encoded data with the hand pose and hand torque as the input to the classifier. 

\begin{figure*}[htbp]
    \centering
    \includegraphics[width=0.95\textwidth]{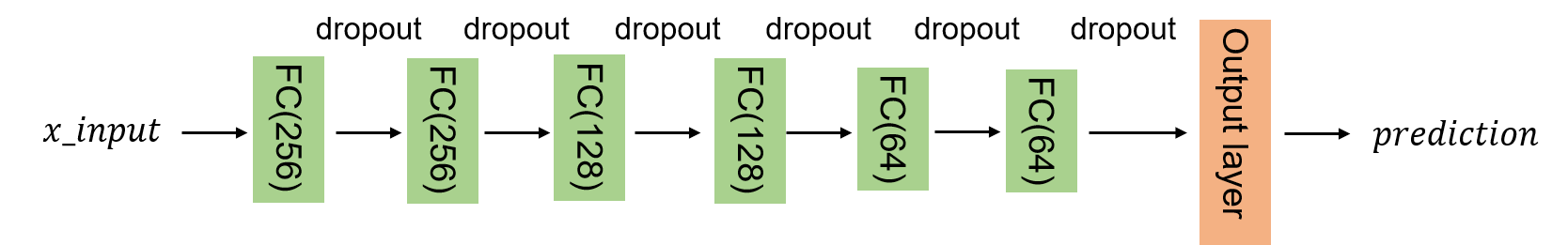}
    \vspace{-3mm}
    \caption{$x$ is the input tactile reading array with the size of $24$ on one of the four regions in Barrett hand, $x\_input$ contains {hand pose, tactile reading (encoded), hand torque} and has a dimension of $34$ and $prediction$ is the classification result. FC(256), FC(128) and FC(64) refers to fully connected layers with $256$, $128$ and $64$ units. The dropout rate is 0.5.}
    \label{fig-classifier}
\end{figure*}

\paragraph{Ensemble classifier}
We train three classifiers - the naive model, encoder model, and encoder-regression model. Their general structures are the same as shown in Figure \ref{fig-classifier}, while their differences are the input data dimension and output layer activation function. The input data dimension of the naive model is $106$, which contains $\{$hand pose, readings from tactile sensors($96$), strain gauge(coupled joints) $\}$, and the activation function of the output layers is softmax. Instead of using the data directly, we use encoded tactile reading as the input to the encoder model, and the input dimension of the encoder model is $34$. The output layer of the encoder model is the same as the naive model. For the encoder-regression model, the input is the same as the encoder model, whereas the output layer is a dense layer with a unit number of 1. We ensemble the output of the three models to represent our final prediction as in Equation \ref{ensemble}.  
\begin{equation} \label{ensemble}
P = \frac{P_{naive} + P_{encoder} + P_{encoder-regression}}{3}
\end{equation}

\paragraph{Transfer learning}
We train the above models using the simulation data and then apply transfer learning to the models with the real system's data since the real system data size is smaller.  We apply transfer learning to the three models separately, then ensemble the models together for prediction.  

\section{Evaluation and Results}
We divide the prediction into $5$ classes: $0$ object, $1$ object, $2$ objects, $3$ objects and more than $4$ objects. The distance between the predicted and the ground truth number is meaningful, therefore, we use the root mean squared error (RMSE) as the metric to evaluate our models' performance. We shall present the results of our baseline models, which are the contact force model and grasp volume model, followed by the result of the ensemble data-driven model. 

\subsection{Baseline model results}
\subsubsection{Estimation using contact force} 
We split the ping-pong ball simulation data into $80\%$ and $20\%$ as training and testing data for the linear regressor, while the ping-pong ball real system data with the percentage of $50\%$ and $50\%$, which is the same ratio used to train the deep learning models. The result is presented in Table \ref{tab-force-result}. N/A (not available) means that case did not occur during testing. We can see that the RMSE after the hand lifts is smaller than the RMSE before the hands lift in both the simulation data and real system data.
This shows that when the hand is within the pile, the tactile force sensor readings could contain noise.
Moreover, the force estimation will not work for more than $3$ ping-pong balls.
Lastly, the estimator output $1$ for all inputs from the real system. This is likely due to noise in the sensor data from the real system.

\begin{table}[h]
\caption{Results of estimating with contact force}
\vspace{-2mm}
\label{tab-force-result}
\begin{center}
\begin{tabular}{|c|c|c|c|c|}
\hline
\textbf{\begin{tabular}[c]{@{}c@{}}estimated \\  object \\ number\end{tabular}} & \textbf{\begin{tabular}[c]{@{}c@{}}before lift \\ (sim) \\ RMSE\end{tabular}} & \textbf{\begin{tabular}[c]{@{}c@{}}after lift \\ (sim) \\ RMSE\end{tabular}} & 
\textbf{\begin{tabular}[c]{@{}c@{}}before lift \\ (real) \\RMSE\end{tabular}} & \textbf{\begin{tabular}[c]{@{}c@{}}after lift \\ (real) \\ RMSE \end{tabular}}\\
\hline
0 & 0.61 & 0.53 & N/A & 0.58\\
\hline
1 & 0.73 & 0.71 & 1.35 & 1\\
\hline
2 & 1.07 & 0.78 & N/A & 0.95\\
\hline
3 & 1.55 & 1.17 & N/A & 0.56\\
\hline
$\geq4$ & N/A & N/A & N/A & N/A\\
\hline
overall & 0.7 & 0.62 & 1.35 & 0.98\\
\hline
\end{tabular}
\end{center}
\end{table}

\subsubsection{Estimation using grasp volume}
We use the real system data of grasping ping-pong balls to test our grasp volume method's performance. The results are shown in Table \ref{tab-volume-result}. The overall RMSE between our estimation and the ground truth is $2.95$. $99.71\%$ of the ground truth grasp numbers were lower or equal to the volume-based grasp number.  Therefore, the upper bound estimation of the number of objects that can be grasped is reliable. Table \ref{tab-volume-result} provides the break-down details. The incorrect upper-bound estimation represents the percentage of estimations that were lower than the ground truth.  

\begin{table}[h]
\caption{Results of estimating with grasp volume}
\vspace{-2mm}
\label{tab-volume-result}
\begin{center}
\begin{tabular}{|c|c|c|}
\hline
\textbf{\begin{tabular}[c]{@{}c@{}}estimated \\  grasp \\ number\end{tabular}} & \textbf{\begin{tabular}[c]{@{}c@{}}ping-pong \\ ball(real) \\  RMSE\end{tabular}} &
\textbf{\begin{tabular}[c]{@{}c@{}}incorrect \\ upper-bound \\estimation (\%)\end{tabular}}\\
\hline
0 & N/A  & N/A\\
\hline
1 & N/A  & N/A\\
\hline
2 & 0.53 & 0\%\\
\hline
3 & 1.77 & 0.41\%\\
\hline
$\geq4$ & 3.25 & 0.26\%\\
\hline
overall & 2.95 & 0.29\%\\
\hline
\end{tabular}
\end{center}
\vspace{-5mm}
\end{table}

\begin{figure*}[htbp]
    \centering
    \makebox[\textwidth][c]{\includegraphics[width=1\textwidth]{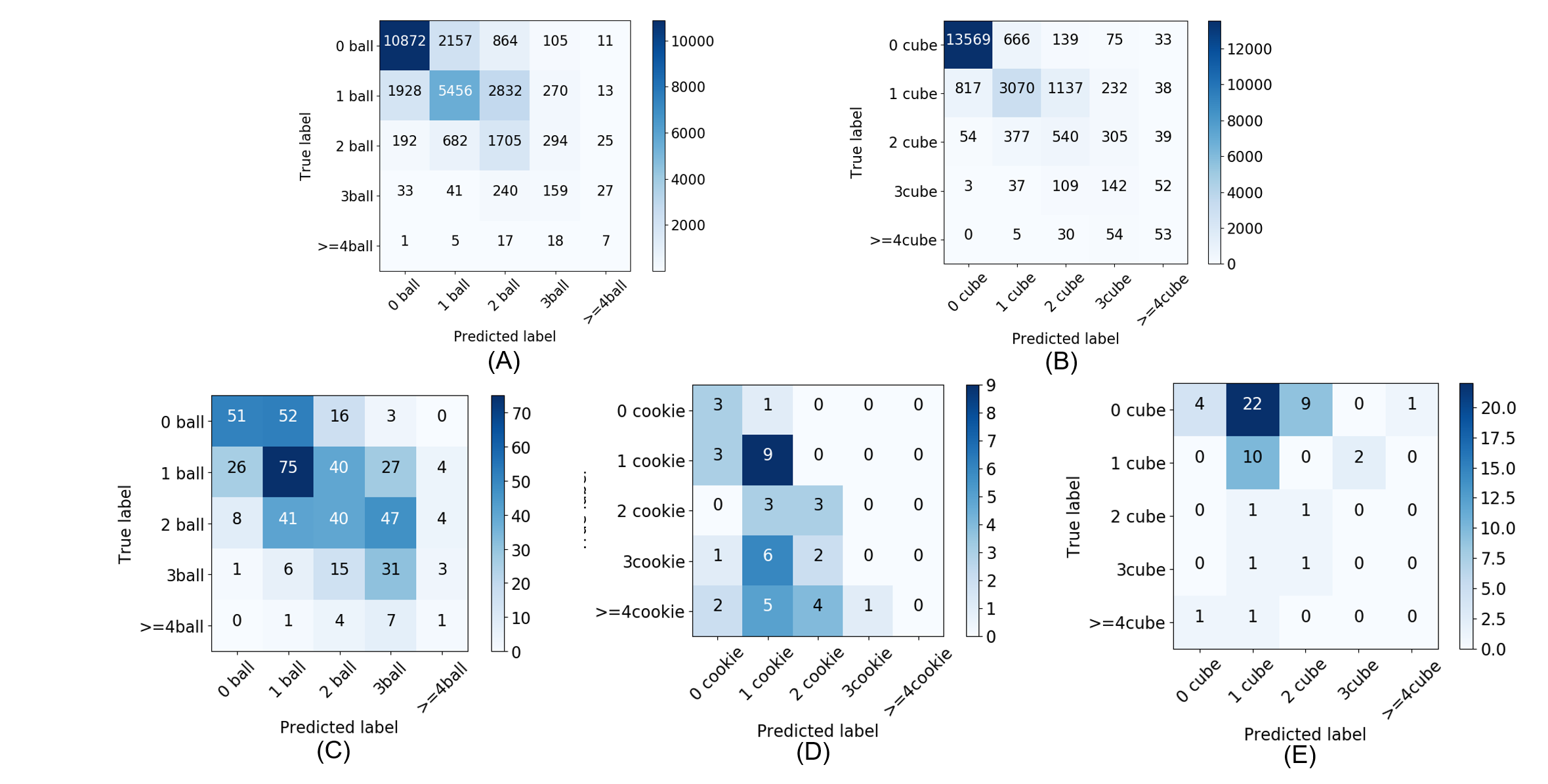}}
    \caption{Confusion matrices of the ensembled models. (A) Confusion matrix for the simulation ping-pong ball data. (B) Confusion matrix for the simulation cube data. (C) Confusion matrix for the real system ping-pong ball data. (D) Confusion matrix for the fortune cookie data. (E) Confusion matrix for the real system foam cube data.}
    \label{fig-confusion_matrix}
\end{figure*}

\subsection{Deep learning model}
\subsubsection{Training the classifiers}
We divide the simulation data into training, validation, and testing with the percentage of $60\%$, $20\%$, and $20\%$. For the encoders training, we use the Adam optimizer with a learning rate of $0.001$, loss function as mean squared error, and epoch number of $3000$. For the classification model's training, we use the Adam optimizer with a learning rate of $0.001$ and an epoch number of $3000$ as well. We use categorical cross-entropy as the loss function for the naive model and encoder model, and use mean squared error as the loss function for the encoder-regression model. The batch size we use is $500$. We also apply oversampling based on each class's weight during the training due to the difficulty of getting data trials with a larger numbers of objects. We train different models on simulation ping-pong ball data and simulation cube data. We apply transfer learning of ping-pong ball and fortune cookie data using pre-trained weights from the simulation ping-pong ball model while foam cube data using pre-trained weights from the simulation cube data. We split the data into training, validation, and testing with the percentage of $40\%$, $10\%$, and $50\%$.  For training the ping-pong ball model in transfer learning, we choose epoch number as $3000$ while for the other two objects, we choose epoch number to be $500$.     

\subsubsection{Results of data-driven approach}

The results of the data-driven models are presented in Table \ref{tab-model-result}. We calculate the RMSE between the prediction of our ensemble model and the ground truth value. The models achieve an overall RMSE of $0.74$ on the ping-pong ball simulation data and $0.58$ on the cube simulation data. The models' performance on real system data through transfer learning achieves an overall RMSE of $1.06$, $1.78$, and $1.45$ for ping-pong balls, fortune cookies, and foam cube, respectively. 

Compared with the performance on the ping-pong ball ensemble model trained with simulation data, the transfer learning model's performance on the real system data achieves a larger RMSE on $0$ object class and $1$ object class. The transfer learning model achieves a smaller RMSE on both $3$ object and $\geq 4$ object classes. The above results lead to an overall larger RMSE of $0.32$ for the transfer learning model than the model trained with the simulation data. 
The transfer learning model of foam cube achieves a larger RMSE in all the cases and results in a larger overall RMSE of $0.87$. The transfer learning results of the ping-pong ball and foam cube are within our expectations since the real system has noisier tactile sensor readings than the simulator's tactile sensor readings. Additionally, the resolution of the tactile sensor readings is the same as the weight of the ping-pong ball and foam cube in the real system. Therefore, even when there is no object in the Barrett hand, there can be similar tactile readings as if objects are grasped by the hand. 

We perform transfer learning for the fortune cookie directly using the pre-trained weights from the ping-pong ball simulation model since it is very hard to design shapes similar to fortune cookies in the simulator. With no surprise, the model performance on fortune cookie is the worst among all the three transfer learning models, which has an overall RMSE of $1.78$. Moreover, the fortune cookie transfer learning model can not predict cases where there are more than $4$ objects in the Barrett hand. This is mainly because the fortune cookie's plastic bags can get stuck into the space between the finger joints on Barrett hand without contributing anything to the tactile sensor readings.

\begin{table}[h]
\caption{Results of estimating with ensembled model}
\label{tab-model-result}
\begin{center}
\begin{tabular}{|c|c|c|c|c|c|}
\hline
\textbf{\begin{tabular}[c]{@{}c@{}}estimated \\ object \\ number\end{tabular}} & \textbf{\begin{tabular}[c]{@{}c@{}}ping-pong \\ ball \\ (sim)\end{tabular}} & \textbf{\begin{tabular}[c]{@{}c@{}}ping-pong \\ ball \\ (real)\end{tabular}} & \textbf{\begin{tabular}[c]{@{}c@{}}fortune \\ cookie \\ (real)\end{tabular}} & \textbf{\begin{tabular}[c]{@{}c@{}}cube \\ (sim)\end{tabular}} & \textbf{\begin{tabular}[c]{@{}c@{}}foam cube \\ (real)\end{tabular}} \\
\hline
0 & 0.48 & 0.88 & 2.21 & 0.27 & 1.79\\
\hline
1 & 0.6 & 0.85 & 1.74 & 0.55 & 1.01\\
\hline
2 & 1.08 & 1.09 & 1.41 & 0.99 & 1.83\\
\hline
3 & 1.66 & 1.27 & 1 & 1.56 & 2\\
\hline
$\geq$4 & 2.25 & 2.14 & N/A & 2.24 & 4\\
\hline
overall & 0.74 & 1.06 & 1.78 & 0.58 & 1.45\\
\hline
\end{tabular}
\end{center}
\end{table}

The confusion matrix for each of the five models can be found in Figure \ref{fig-confusion_matrix}. The overall accuracy for the models from (A) to (E) is $65.1\%$, $80.52\%$, $39.36\%$, $34.88\%$ and $27.78\%$ respectively. In general, the models can predict $0$ and $1$ object cases better than the others. When the ground truth object number is $2$, the models confuse $2$ with $1$ or $3$ objects. When the number of objects in the robot hand is larger than $2$, the models struggle to estimate the number of objects grasped.
This is because when the number of objects grasped increases, objects will likely lie on one another without touching the tactile sensors or have minimal contact with the tactile sensors. Thus, $2$, $3$, and $\geq4$ cases can have similar tactile readings at the moment just before the hand lifts. Another phenomenon we noticed is that in the transfer learning models trained with real system data, the models' performance will get worse on the $0$ object case. Instead of classifying the $0$ cases as $0$, they are preferred to be classified as $1$ or more. This implies that the real system's noise is relatively high compared to the weight of the objects.

\section{Conclusion and Discussion} 
We have studied an unexplored robotic grasping problem -- multi-object grasping.  In this problem, accurately estimating the number of objects in a grasp is critical.  We present three approaches that can estimate the number of objects in a grasp and explore their advantages and limitations. 

\begin{figure}[htbp]
    \centering
    \includegraphics[width=0.7\linewidth]{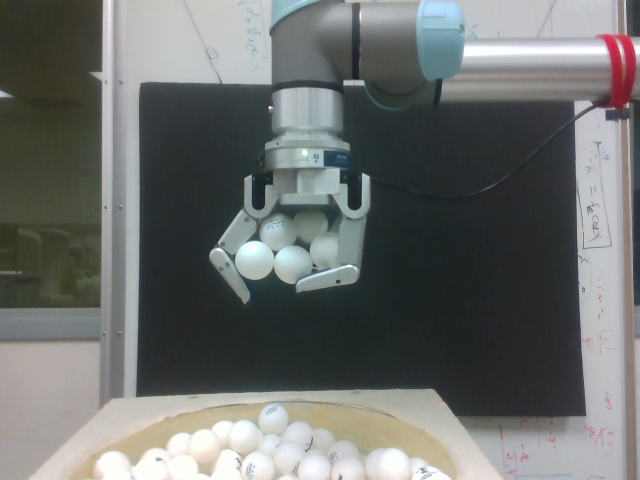}
    \caption{Example of 5 ping-pong balls are in the grasp}
    \label{fig-5_ping-pongball}
\end{figure}

From the experimental results, we can see that the contact force model gives the lowest RMSE error on the simulation data but is very unstable in the real system. The grasp volume approach calculates the volume of a grasp based on the hand's posture. It can estimate the maximum number of objects in a grasp reasonably accurately, however, performs poorly when estimating the exact number. The data-driven model can provide good estimations dealing with both simulation and real system data. However, the estimation accuracy of the data-driven model drops rapidly when the number of objects grasped increases. We think that the main reason is that when there are more than $2$ objects grasped by the hand, they may not all be in contact with the tactile sensors, and the readings can be very similar to grasping fewer objects. Figure \ref{fig-5_ping-pongball} provides one example. The rightmost ping-pong ball has a minimal contact area with the tactile sensors, and the left pint-pong ball on the top level is not in contact with any tactile sensor.

\section*{Acknowledgment}
This material is based upon work supported by the National Science Foundation under Grants Nos. 1560761, 1812933, and 191004.

\bibliographystyle{IEEEtran}
\bibliography{references}

\end{document}